\documentclass{article}

\usepackage{PRIMEarxiv}

\usepackage[utf8]{inputenc} 
\usepackage[T1]{fontenc}    
\usepackage{hyperref}       
\usepackage{url}            
\usepackage{booktabs}       
\usepackage{amsfonts}       
\usepackage{nicefrac}       
\usepackage{microtype}      
\usepackage{lipsum}
\usepackage{fancyhdr}       
\usepackage{graphicx}       
\graphicspath{{media/}}     
\usepackage[numbers]{natbib}
\usepackage{doi}
\usepackage{tabularx}
 
\usepackage{multirow}
\usepackage{array, booktabs, makecell}
\usepackage{siunitx, mhchem}
 
\usepackage{xspace}

 \usepackage{amsmath,amssymb,amsfonts}
 \usepackage{algorithmic}

\usepackage{array, booktabs, makecell}
\usepackage{siunitx, mhchem}
\usepackage{textcomp}
\usepackage{xcolor}

\title{Enhancing Object Detection Robustness: A Synthetic and Natural Perturbation Approach
 
}

\author{Nilantha Premakumara\textsuperscript{\textasteriskcentered}, Brian Jalaian\textsuperscript{\textdagger {\textasteriskcentered}}, Niranjan Suri\textsuperscript{\textasteriskcentered}, and Hooman Samani\textsuperscript{\textdaggerdbl} \\
\textsuperscript{\textasteriskcentered}Institute for Human \& Machine Cognition (IHMC), Pensacola, Florida, USA \\
\textsuperscript{\textdagger}University of West Florida, Pensacola, Florida, USA \\
\textsuperscript{\textdaggerdbl}University of Hertfordshire, Hatfield, UK \\
\{npremakumara, bjalaian, nsuri\}@ihmc.org\\
\textsuperscript{\textdagger}bjalaian@uwf.edu
\textsuperscript{\textdaggerdbl}h.samani@herts.ac.uk}

\begin{document}
\maketitle

\begin{abstract}
 Robustness against real-world distribution shifts is crucial for the successful deployment of object detection models in practical applications. In this paper, we address the problem of assessing and enhancing the robustness of object detection models against natural perturbations, such as varying lighting conditions, blur, and brightness. We analyze four state-of-the-art deep neural network models, Detr-ResNet-101, Detr-ResNet-50, YOLOv4, and YOLOv4-tiny, using the COCO 2017 dataset and ExDark dataset. By simulating synthetic perturbations with the AugLy package, we systematically explore the optimal level of synthetic perturbation required to improve the models' robustness through data augmentation techniques. Our comprehensive ablation study meticulously evaluates the impact of synthetic perturbations on object detection models' performance against real-world distribution shifts, establishing a tangible connection between synthetic augmentation and real-world robustness. Our findings not only substantiate the effectiveness of synthetic perturbations in improving model robustness, but also provide valuable insights for researchers and practitioners in developing more robust and reliable object detection models tailored for real-world applications.
\end{abstract}


\section{Introduction}
Object detection, a fundamental and critical problem in computer vision, aims to identify and spatially localize objects within images or videos \cite{Felzenszwalb2010}. It is integral to various computer vision tasks, including object tracking \cite{Kang2018}, activity recognition \cite{1211526}, image captioning \cite{Wu2018}, image segmentation \cite{Hariharan2014}, and visual question answering \cite{Alberti2019}. Object detection poses a significant challenge due to high intra-class and low inter-class variance \cite{Everingham2010}.

In recent years, deep learning has achieved unprecedented advances in various domains, such as generative modeling \cite{Zhu2017}, computer vision \cite{Sabour2017}, and natural language processing \cite{Devlin2018}. As deep learning techniques are incorporated into safety-critical application areas like autonomous vehicles, medical diagnostics, and robotics, ensuring the reliability and trustworthiness of these systems is paramount.

Although object detection models have demonstrated remarkable performance, real-world applications do not guarantee high-quality input images. Therefore, evaluating the robustness of these models against distribution shifts is essential before deployment. Numerous robustness interventions have been proposed to improve neural network robustness, including non-conventional architectures\cite{Vasconcelos2020}, adding new images to training data \cite{Zhang2017}, alternative losses \cite{Kornblith2020}, and different optimizers \cite{Metz2019}. These interventions target specific distribution shifts, such as noises \cite{Zhou2017} or synthetic corruptions \cite{Hendrycks2019}.

Our work presents a novel approach to assessing the robustness of object detection models by comparing their performance against synthetic and real natural perturbations. We evaluate four state-of-the-art deep neural network models, Detr-ResNet-101, Detr-ResNet-50, YOLOv4, and YOLOv4-tiny, using the COCO 2017 dataset and the AugLy augmentation package, which provides synthetic approximations of natural perturbations. Subsequently, we conduct a comprehensive ablation study to perform transfer learning by re-training these models on synthetic natural perturbations and evaluating their robustness against real perturbations using the ExDark dataset.

The primary contributions of this study offer a novel perspective on enhancing the robustness of object detection models against real-world distribution shifts. We meticulously unravel the intricate relationship between synthetic and real-world perturbations through the following innovative approaches:
\begin{itemize}
\item We delve into a systematic exploration to experimentally identify the optimal level of synthetic perturbation that effectively enhances the selected models' robustness, shedding light on the potential benefits of synthetic data augmentation for real-world deployment.
\item Our comprehensive ablation study meticulously evaluates the impact of synthetic perturbations on object detection models' performance against real-world distribution shifts, establishing a tangible connection between synthetic augmentation and real-world robustness.
\end{itemize}
Our findings not only substantiate the effectiveness of synthetic perturbations in improving model robustness, but also provide valuable insights for researchers and practitioners in developing more robust and reliable object detection models tailored for real-world applications.

\section{Related Work}
 The robustness of CNN-based object detection techniques in the presence of noise is a critical area of research in computer vision, especially for surveillance applications where image quality can be challenging. Several studies have investigated the impact of noise on object detection algorithms, exploring the effects of synthetic rain, fog, and other degradations on image quality. \cite{Dodge2016} found that noisy images significantly affect classification tasks, while \cite{Volk2019} demonstrated that synthetic rain impacts object detection algorithms similarly to real rain.

\cite{Bernuth2019} proposed a novel approach of adding realistic and varying snow and fog to existing image datasets to investigate their potential for scene reconstruction. \cite{Shankar2019} evaluated various classifiers trained on ImageNet and found a median 16\% decline in classification accuracy when exposed to natural perturbations. They also discovered that natural perturbations cause classification and localization errors, resulting in reduced detection mAP for Faster R-CNN and R-FCN models. \cite{Kristo2020} assessed the effectiveness of popular deep learning techniques for object detection and recognition in thermal surveillance scenarios using a custom dataset collected in different weather conditions at night.

\cite{Shetty2020} suggested semantic adversarial editing as a technique to synthesize believable corruptions and highlighted the challenging data points on which their target model should be robust. They showed that their proposed technique can generate a wide range of corruptions and improve model robustness against natural corruptions.
The performance of deep neural network-based methods is directly influenced by the availability and quality of datasets. Several challenging large-scale datasets have been introduced, consisting of images or videos captured under adverse conditions. These datasets include ExDARK, UNIRI-TID, RESIDE, UFDD, and See in the Dark \cite{Ahmed2021}. Each dataset focuses on different aspects of challenging conditions, such as low light, various weather conditions, and occlusions. Table \ref{Literature} summarizes the main advantages and limitations of the mentioned approaches.
\begin{table}[ht]
\label{table:Literature}
\caption{A summary of advantages and limitations of methods tackling object detection in challenging conditions.} 
\label{Literature}
\centering
\begin{tabularx}{\linewidth}{l|X|X|X}
\toprule
Literature & Methods & Advantages &  Limitations \\
\midrule
\cite{9134757} & Image are transformed and then fed into the RFB-Net. & Context information fusion allows detection of object in low-light. & Relies on prior information about type of object, shape etc for detecting them in night-time. \\  
\hline
\cite{inproceedings} & Fusion of pre-trained models using Glue layer and information distillation. & Domain joining with help of glue layer reduces in computation and provides more information for models to learn from different domains. & Relies on prior domain knowledge. \\
\hline
\cite{9235491} & Generative adversarial network with Faster R-CNN. & Networks learns both day and night-time features. & Relies on prior information of converting night-time images to day time. \\
\hline
\cite{wang2019fast} &  Fully convolutional siamese networks with modified binary segmentation task. & Pre-frame binary segementation mask is used for low-level object representation instead of relying on feature extractor backbone. & Relies on prior information while generating binary segmentation mask. Fails when faced with motion blur and non-object pattern. \\
\hline
\cite{tu2022rgbt} & Two stream convolutional neural network with attention mechanism. & Fusion of rgb and thermal image to generate features and noise reduction with convolutional block attention modele. & Thermal images are not efficient for every envirment use. \\
\bottomrule
\end{tabularx}
\end{table}
Recent studies have demonstrated that data augmentation can enhance model resilience \cite{Robert2019},\cite{Hendrycks2019}. Various augmentations can aid resilience, including different types of noise \cite{Lopes2019}, deep artificial image transformations \cite{Zhang2017}, natural transformations \cite{Laugros2021}, and combinations of simple image transformations such as Python Imaging Library operations \cite{Cubuk2018},\cite{Hendrycks2019}. In this research, we used AugLy for image perturbation. AugLy is a state-of-the-art open-source Python library designed to help AI researchers apply data augmentations to assess and improve the robustness of their machine learning models. It integrates multiple modalities, including image, video, and text, which is essential in many AI research domains \cite{Zoe2022}. AugLy provides over 100 data augmentations that mimic what real users do to photos and videos on platforms like Facebook and Instagram. For our experiment, we used three image augmentations: blur, pixel degradation, and brightness.

In summary, the related work underscores the importance of developing robust deep learning models capable of handling natural data variance and corruptions. Researchers have explored various methods, such as generative models, data augmentation, and synthetic corruption benchmarks, to achieve this goal. The availability of challenging datasets and the application of data augmentation techniques, such as those provided by AugLy, play a crucial role in assessing and enhancing model robustness under adverse conditions. Our research builds upon these findings and aims to further investigate the impact of synthetic natural perturbations on model performance and the potential benefits of transfer learning and retraining on synthetic perturbations to improve robustness against real perturbations.

\section{Methodology}
\label{sec:methodology}
\subsection{Models, Dataset, and Perturbations}
\begin{figure}[h]
\centering
\includegraphics[scale=0.40]{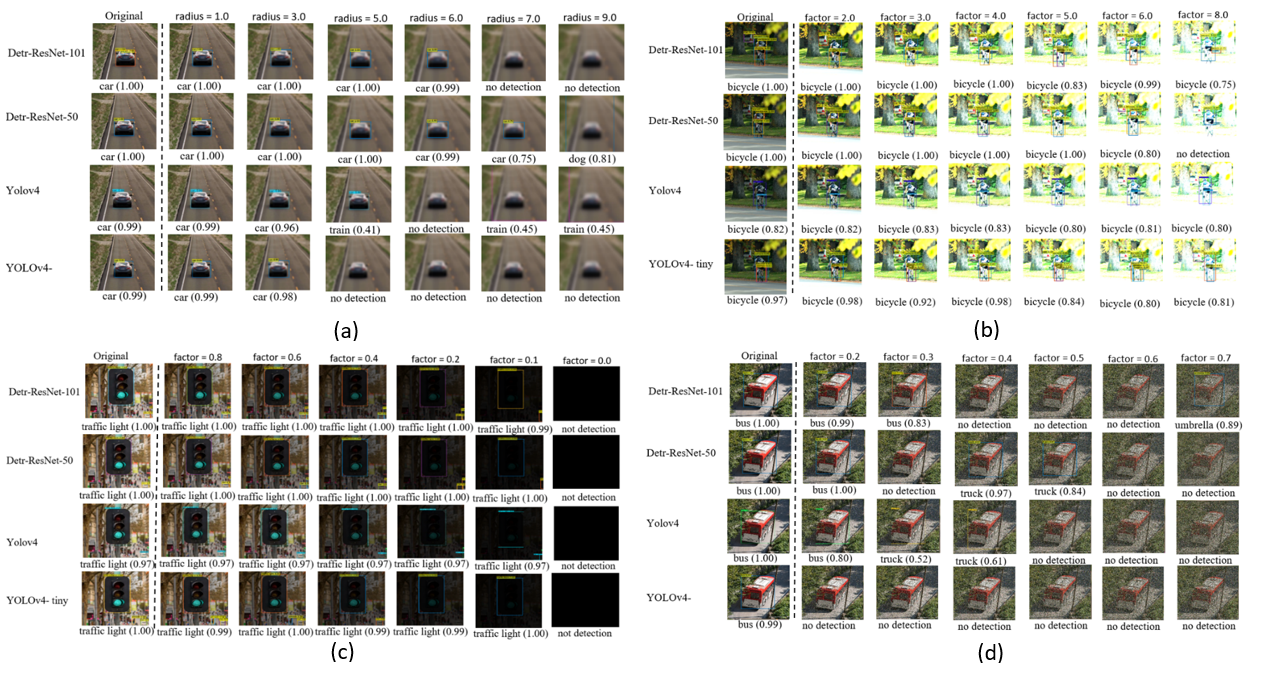}
\caption{Illustration of how each model is robust against perturbation level (a) blur, (b) brightness (light mood), (c) brightness (dark mood), (d) degrading pixels.}
\label{fig:S-Example}
\end{figure}

We evaluated model robustness using four pretrained state-of-art neural network models, a test dataset, and synthetic perturbations. The main components of our methodology are as follows:

\begin{itemize}
\item \textbf{Models:} We used four pretrained neural network models: Detr-ResNet-101, Detr-ResNet-50, Yolov4, and Yolov4-tiny.

\item \textbf{Dataset:} The COCO 2017 dataset was employed for training purposes and for creating synthetic perturbations. The evaluation of the models' robustness and performance was performed using the ExDARK dataset \cite{Exdark}, containing images with real-world perturbations, as part of our ablation study.

\item \textbf{Synthetic Perturbations:} We introduced three types of robustness specifications (blur, brightness, and pixel degradation) using the AugLy package. Perturbation parameters were adjusted within specific ranges to simulate natural perturbations. To simulate model robustness with natural distribution shifts on the image data, we adopted the perturbation strategies on the dataset. We used perturbations with different levels for each category. Figure \ref{fig:S-Example} depicts how each model's confidence score corresponds to the perturbation level. Example results show that all the model's detection confidence scores decrease when increasing perturbation level compared to the original one.
\end{itemize}

\subsection{Ablation Study Design}
Ablation studies are experimental approaches used to assess the importance of individual components in complex systems or models, such as deep learning models, by removing or altering these components \cite{autoablation}. This methodology has been widely adopted in computer vision, particularly for object detection tasks \cite{rcnn}, and in other domains such as natural language processing and neuroscience \cite{meyes2019ablation}.

In this paper, we conduct an ablation study using all four models to evaluate the effectiveness of synthetic perturbations, specifically focusing on the brightness modality, in enhancing the robustness of object detection models against real-world distribution shifts. We test with varied numbers of synthetic perturbation images to determine how the size of the synthetic perturbation training set influences detection performance on natural perturbations. Our ablation study serves two primary purposes:

\begin{enumerate}
\item \textbf{Quantify the impact of synthetic perturbations:} We apply varying levels of synthetic brightness perturbations to identify the optimal level that improves the models' robustness. This process helps us understand the relationship between the perturbation magnitude and the performance of the object detection models.

\item \textbf{Analyze the transferability of improvements:} We investigate whether the models' robustness improvements due to synthetic brightness perturbations translate to enhanced performance against real-world distribution shifts, thereby determining if synthetic perturbations serve as reliable proxies for natural perturbations in real-world applications.
\end{enumerate}

Our findings from the ablation study provide valuable insights into developing more robust object detection models and inform future research on the role of synthetic perturbations in enhancing model resilience. 

\section{Results}
\subsection{Model Performance with Synthetic Perturbations}

We first evaluated the models' performance against synthetic natural perturbations created using the AugLy augmentation package. Table \ref{mAPValues} shows the mean average precision (mAP) scores of the four object detection models when tested with these synthetic perturbations. Most models were susceptible to strong brightness, while all models were more robust to darkness than other perturbations.
 \begin{table}[htbp]
\label{table:mAPValues}
\caption{The mean average precision (mAP) scores of various models computed with the original dataset and synthetic corruptions}
\label{mAPValues}
\resizebox{\columnwidth}{!}{%
\centering  
  \begin{tabular}{l|l|S|S|S|S|S|S|S|S}
    \toprule
    \multirow{2}{*}{Model} &  
    \multirow{2}{*}{\thead{mAP\\Original}} &   
      \multicolumn{2}{c|}{Blur(radius=5)} &
      \multicolumn{2}{c|}{\thead{Brightness\\(factor=3)}} &
      \multicolumn{2}{c|}{\thead{Brightness\\(factor=0.2)}} &
      \multicolumn{2}{c}{\thead{Degrading pixels\\(radius=5)}} \\
      & & {mAP} & {mAP$\Delta$} & {mAP} & {mAP$\Delta$} & {mAP} & {mAP$\Delta$} & {mAP} & {mAP$\Delta$}\\
      \midrule
   Detr-ResNet-101 & 73.56 & 67.26 & 6.3 & 64.39 & 9.17 & 69.02 & 4.54 & 64.53 & 9.03\\
   Detr-ResNet-50 & 62.37 &51.26 & 11.11 & 47.89 & 14.48 & 59.21 & 3.16 & 48.52 & 13.85\\
   Yolov4 & 71.49 & 63.37 & 8.12 & 62.44 & 8.12 & 68.21 & 3.28 & 61.85 & 9.64\\
   Yolov4-tiny& 64.52 & 53.47 & 11.05 & 51.04 & 11.05 & 60.83 &3.69& 50.25 & 14.27 \\
    \bottomrule
  \end{tabular}
  }
\end{table}
\subsection{Comparison of Model Robustness}
\begin{figure}[htbp]
\centering
\includegraphics[scale=0.35]{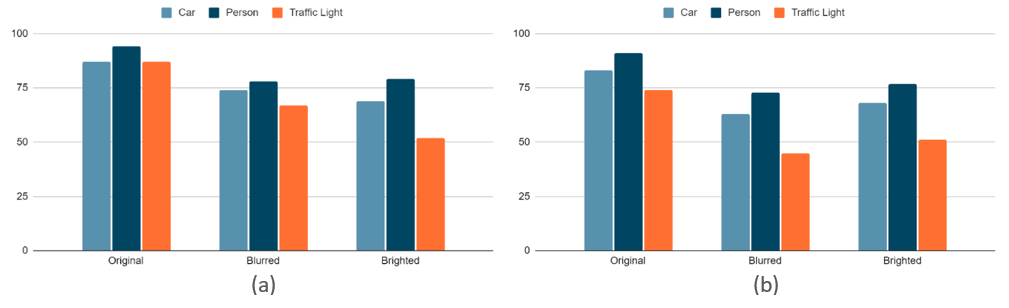}
\caption{Illustration of performance of multiple object detection models against AugLy blur and brightness perturbation (a) Detr-ResNet-101 model and (b) Detr-ResNet-50 model}
\label{fig:comparison}
\end{figure}

To provide a comprehensive comparison of the robustness of the four object detection models (Detr-ResNet-101, Detr-ResNet-50, YOLOv4, and YOLOv4-tiny), we analyzed their performance under different perturbation conditions. As shown in Figure \ref{fig:comparison}, Detr-ResNet-101 and Detr-ResNet-50 exhibit similar performance patterns under various perturbations, with Detr-ResNet-101 performing slightly better than Detr-ResNet-50. YOLOv4 outperforms YOLOv4-tiny in all scenarios.

Overall, the Detr-ResNet-101 model demonstrated the best performance in handling synthetic perturbations and real-world perturbations when trained with an augmented synthetically perturbed dataset. However, it should be noted that the choice of the best model for specific scenarios may depend on various factors, including the computational power available, the acceptable trade-off between accuracy and speed, and the specific perturbation types encountered in the target domain.
\subsection{Impact of Perturbation Levels}
To investigate the models' performance against varying levels of synthetic perturbation, we conducted experiments by changing the radius and factor values of each synthetic perturbation. This experiment aimed to identify the optimal level of synthetic perturbation that should be used to improve the model's robustness.

\begin{figure}[h]
    \centering
    \includegraphics[scale=0.40]{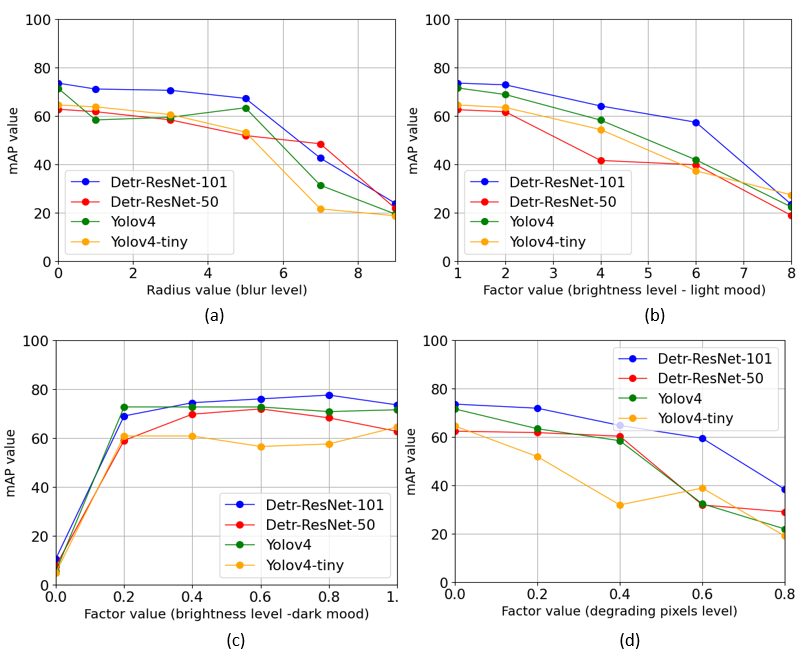} 
	\caption{Illustration of model's performance against different synthetic perturbation levels.(a),(b),(c),(d) are represent different synthetic blur level, brightness (light mood) level,  synthetic brightness (dark mood) level and synthetic degrading pixels level respectively}
	\label{fig:model's performance-level}
\end{figure}

In Figure \ref{fig:model's performance-level}, we observed that the models' performance worsened when the blur radius value was greater than $5$, the brightness (light mood) factor value was greater than $2$, the brightness (dark mood) factor value was less than $0.2$, and the degrading pixels factor value was greater than $0.2$. Among the models, Detr-ResNet-101 demonstrated higher robustness compared to the other three models.
\subsection{Ablation Study: Synthetic to Real Perturbation Robustness}
In the ablation study, we examined the performance of the four object detection models (Detr-ResNet-101, Detr-ResNet-50, YOLOv4, and YOLOv4-tiny) against real perturbations from the ExDark dataset, which contains challenging lighting conditions from an out-of-domain setting. The models were re-trained with synthetic perturbations before testing. We conducted further tests with varied numbers of synthetic perturbation images to determine how the size of the synthetic perturbation training set influences detection performance on real perturbations. To test the outcomes, we added synthetic poor brightness perturbations to a subset of the COCO-2017 training dataset at random percentages of 0\%, 20\%, 50\%, and 70\% using the AugLy package.

\begin{figure}[htbp]
    \centering
    \includegraphics[scale=0.35]{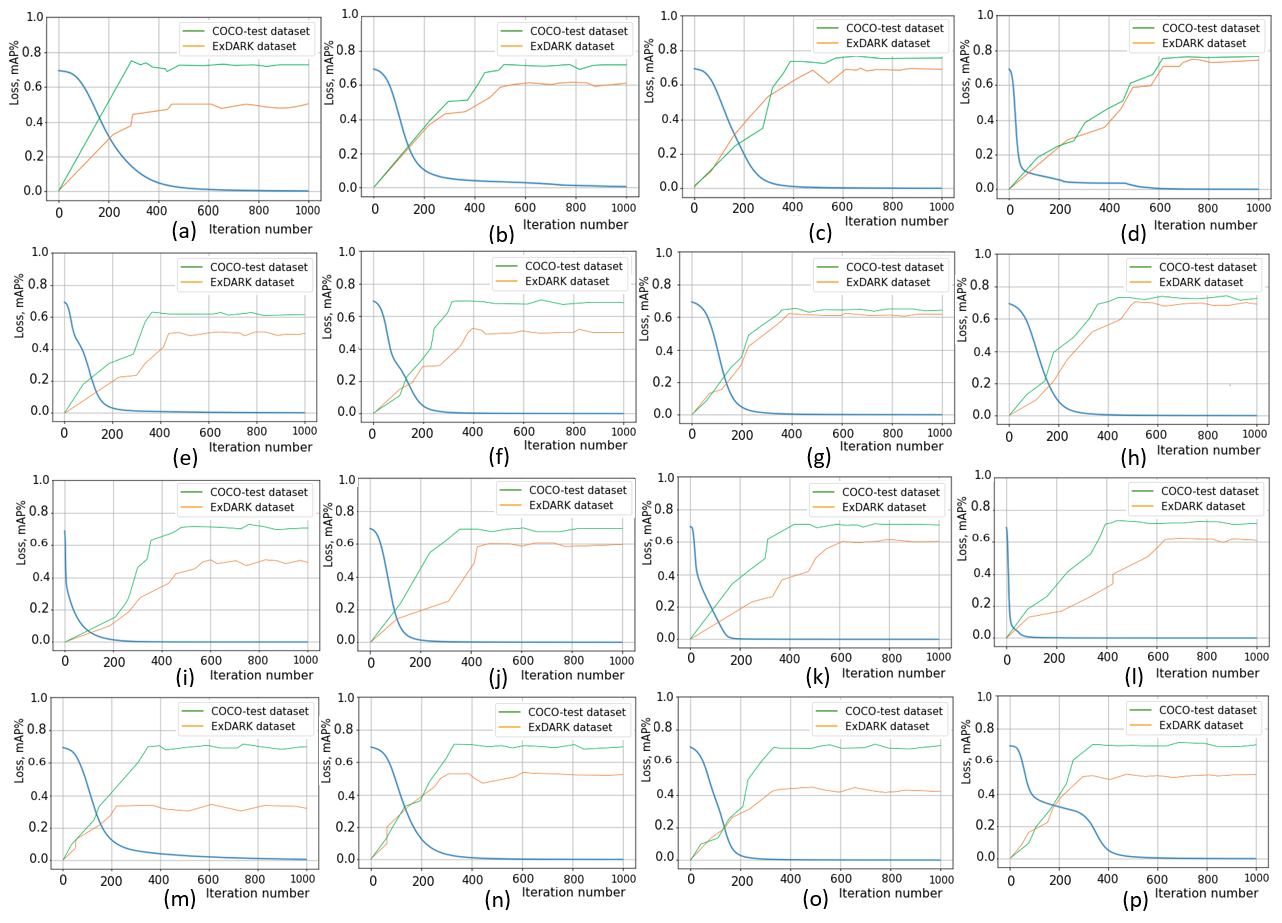} 
	\caption{mAP and loss function graphs for different models and synthetic ratio combinations. (a-d) Detr-ResNet-101, (e-h) Detr-ResNet-50, (i-l) YOLOv4, and (m-p) YOLOv4-tiny. Columns represent original to synthetic ratios of 100:0, 80:20, 50:50, and 30:70, respectively.}
	\label{fig:all-model's performance-new}
\end{figure}

Figure \ref{fig:all-model's performance-new} shows the values of the mAP and loss functions for the models with subset ratios of 100:00, 80:20, 50:50, and 30:70, respectively. Our results demonstrate that the mAP has an increasing trend as we increase the percentage of the augmented synthetically perturbed training dataset and re-train the models. The common observation across all cases is that while there is a significant gap between the mAP for the original models, the gap narrows as the percentage of the augmented synthetically perturbed training dataset increases. This suggests that adding synthetic perturbations to the training data improves the models' robustness to real-world perturbations. However, in this study, we focused on the results with brightness adjustments only, as presented in Table \ref{Ablationstudy}.

\begin{table}[ht]
\label{table:Ablationstudy}
\caption{The mAP scores of various models computed with the original dataset and synthetic corruptions}
\label{Ablationstudy}
\resizebox{\columnwidth}{!}{%
\centering
  \begin{tabular}{c|c|S|S}
    \toprule
    \multirow{2}{*}{Model} &  
    \multirow{2}{*}{\thead{Original:Synthetic\\ratio on traing dataset}} &   
      \multicolumn{2}{c}{Model's performance (mAP)} \\
      & & {COCO-testing} & {ExDARK}\\
      \midrule
   Detr-ResNet-101 & 100:00 & 73.56 & 51.85 \\
     & 80:20 & 73.24 & 61.43 \\
     & 50:50 & 75.23 & 70.56 \\
    & 30:70 & 77.92 & 76.47\\
    \hline
    Detr-ResNet-50 & 100:00 & 62.37 & 48.28 \\
     & 80:20 & 64.76 & 58.59 \\
     & 50:50 & 63.47 & 62.58 \\
    & 30:70 & 67.58 & 66.89\\
    \hline
    Yolov4 & 100:00 & 71.49 & 49.48 \\
     & 80:20 & 68.78 & 60.45 \\
     & 50:50 & 70.37 & 61.57 \\
    & 30:70 & 72.45 & 62.45\\
     \hline
    Yolov4-tiny & 100:00 & 64.52 & 32.45 \\
     & 80:20 & 65.43 & 38.58 \\
     & 50:50 & 67.84 & 42.59 \\
    & 30:70 & 69.51 & 56.73\\
    \bottomrule
  \end{tabular}
  }
\end{table}

\section{Limitations and Future Work}

While our study provides valuable insights into the performance of object detection models under various synthetic perturbations and their robustness against real-world perturbations, there are several limitations that should be acknowledged. First, our analysis was limited to the ExDark dataset, which primarily focuses on poor lighting conditions. To generalize our findings, it is crucial to create and test the models on datasets containing different modalities of natural perturbations, such as occlusions, weather variations, and sensor noise, among others.

Second, our experiments only considered four pre-trained object detection models. To provide a more comprehensive understanding of model robustness, future work could include a wider range of models and architectures, as well as customized models specifically designed to handle perturbations.

Another limitation is that our study focused mainly on brightness adjustments in the ablation study. Expanding the scope of the ablation study to include other types of synthetic perturbations could provide a more complete understanding of model performance and generalization.

Lastly, the choice of augmentation techniques and their parameter settings may also impact the results. Future work could explore alternative data augmentation methods and fine-tune the parameters to find the optimal balance between model robustness and performance.

In summary, future work should aim to:
\begin{enumerate}
    \item Develop and test object detection models on diverse datasets containing various natural perturbation modalities.
    \item Investigate the performance and robustness of a broader range of object detection models and architectures.
    \item Expand the ablation study to encompass different types of synthetic perturbations.
    \item Explore alternative data augmentation techniques and optimize their parameters to enhance model robustness and performance.
\end{enumerate}
\section{Conclusion}
In this study, we investigated the robustness of four state-of-the-art object detection models, namely Detr-ResNet-101, Detr-ResNet-50, YOLOv4, and YOLOv4-tiny, to natural perturbations by simulating synthetic perturbations using the AugLy package. We conducted a series of experiments to evaluate the performance of the models against synthetic and real-world perturbations, assess the impact of the size of the synthetic perturbation training set on detection performance under real perturbations, and determine the optimal level of synthetic perturbation for improving model robustness. Ablation studies were also performed to further analyze the models' performance.

Our findings indicate that the performance of object detection models can be significantly improved by augmenting the training dataset with synthetically perturbed images, particularly in scenarios involving challenging lighting conditions, such as those found in the ExDark dataset. Among the models tested, Detr-ResNet-101 exhibited the best robustness to perturbations. However, the choice of the most suitable model may depend on various factors, including computational power, the trade-off between accuracy and speed, and the specific perturbation types encountered in the target domain.

These results provide valuable insights for researchers and practitioners seeking to enhance the robustness of object detection models to natural perturbations and offer a foundation for further research into methods for improving model performance under a wide range of challenging conditions. By addressing the limitations outlined in the previous section and expanding the scope of investigation, future work can contribute to the development of more robust object detection models capable of handling diverse and complex real-world scenarios.

\bibliographystyle{unsrt}  

\end{document}